\documentclass[runningheads]{llncs}
\usepackage{graphicx}
\usepackage{tikz}
\usepackage{comment}
\usepackage{amsmath,amssymb} % define this before the line numbering.
\usepackage{color}
\usepackage{booktabs}
\usepackage{wrapfig}
\usepackage{hyperref}
\usepackage{multirow}
\usepackage[font=small,labelfont=bf]{caption}
\DeclareMathAlphabet{\mathpzc}{T1}{pzc}{m}{it}

\begin{document}
% \renewcommand\thelinenumber{\color[rgb]{0.2,0.5,0.8}\normalfont\sffamily\scriptsize\arabic{linenumber}\color[rgb]{0,0,0}}
% \renewcommand\makeLineNumber {\hss\thelinenumber\ \hspace{6mm} \rlap{\hskip\textwidth\ \hspace{6.5mm}\thelinenumber}}
% \linenumbers
\pagestyle{headings}
\mainmatter
\def\ECCVSubNumber{35}  % Insert your submission number here

\title{End-to-End 3D Multi-Object Tracking and Trajectory Forecasting\vspace{-0.6cm}} % Replace with your title

% INITIAL SUBMISSION 
\begin{comment}
\titlerunning{ECCV-20 submission ID \ECCVSubNumber} 
\authorrunning{ECCV-20 submission ID \ECCVSubNumber} 
\author{Anonymous ECCV submission}
\institute{Paper ID \ECCVSubNumber}
\end{comment}
%******************

% CAMERA READY SUBMISSION
% \begin{comment}
\titlerunning{End-to-End 3D Multi-Object Tracking and Trajectory Forecasting}
% If the paper title is too long for the running head, you can set
% an abbreviated paper title here
%
\author{Xinshuo Weng\thanks{First two authors contributed equally to this work.} \and
        Ye Yuan$^{\star}$ \and 
        Kris Kitani}
\authorrunning{Weng et al.}
% First names are abbreviated in the running head.
% If there are more than two authors, 'et al.' is used.
%
\institute{\vspace{-0.1cm}Robotics Institute, Carnegie Mellon University \\
\email{\{xinshuow, yyuan2, kkitani\}@cs.cmu.edu}\vspace{-0.1cm}}

% \end{comment}
%******************
\maketitle

\begin{abstract}
\vspace{-0.6cm}
3D multi-object tracking (MOT) and trajectory forecasting are two critical components in modern 3D perception systems. We hypothesize that it is beneficial to unify both tasks under one framework to learn a shared feature representation of agent interaction. To evaluate this hypothesis, we propose a unified solution for 3D MOT and trajectory forecasting which also incorporates two additional novel computational units. First, we employ a feature interaction technique by introducing Graph Neural Networks (GNNs) to capture the way in which multiple agents interact with one another. The GNN is able to model complex hierarchical interactions, improve the discriminative feature learning for MOT association, and provide socially-aware context for trajectory forecasting. Second, we use a diversity sampling function to improve the quality and diversity of our forecasted trajectories. The learned sampling function is trained to efficiently extract a variety of outcomes from a generative trajectory distribution and helps avoid the problem of generating many duplicate trajectory samples. We show that our method achieves state-of-the-art performance on the KITTI dataset. Our project website is at \url{http://www.xinshuoweng.com/projects/GNNTrkForecast}.
\vspace{-0.1cm}
\vspace{-0.4cm}
\end{abstract}

\vspace{-0.5cm}
\section{Introduction}
\vspace{-0.2cm}

\begin{figure}[t]
\vspace{-0.0cm}
\begin{center}
\includegraphics[trim=0cm 0cm 5cm 0cm, clip=true, width=0.7\linewidth]{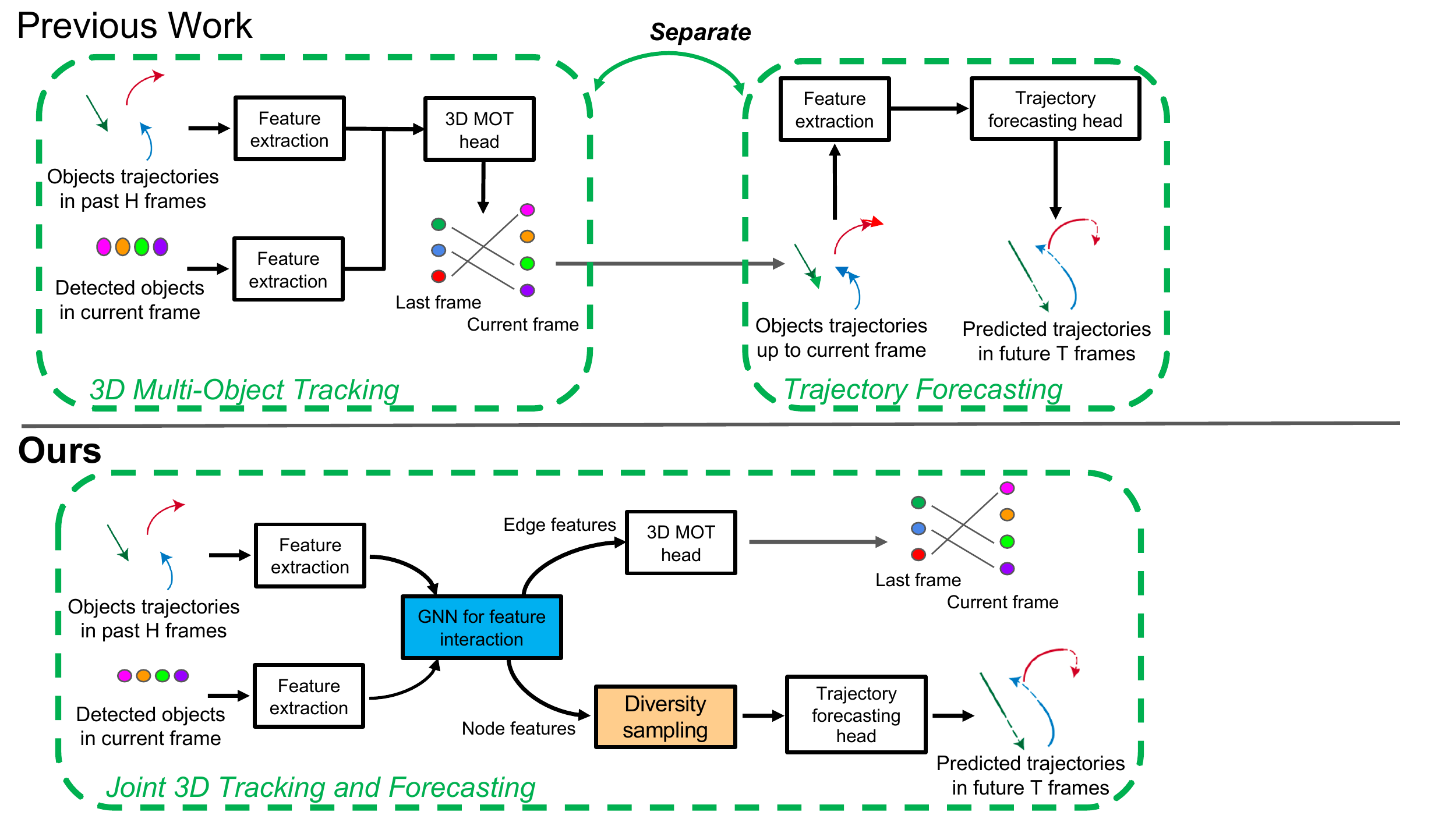}\\
\vspace{-0.35cm}
\caption{\textbf{(Top)} Previous work: 3D MOT and trajectory forecasting treated separately and connected as a sequential process. \textbf{(Bottom)} Our proposed model: Joint process for tracking and forecasting. Two key innovations: (1) a feature interaction using GNNs (\textcolor{blue}{blue box}) to improve tracking association and trajectory forecasting in the presence of multiple agents; (2) a diversity sampling (\textcolor{orange}{orange box}) to improve sample efficiency and produce diverse and accurate trajectory samples.}
\label{fig:teaser}
\vspace{-0.9cm}
\end{center}
\end{figure}

% ==================================================================================================================

% Motivation for joint tracking and forecasting
3D multi-object tracking (MOT) and trajectory forecasting are critical components in modern perception systems. Historically, MOT \cite{Weng2019,Wang2020_GNNDetTrk,Weng2020_GNN3DMOT} and trajectory forecasting \cite{Gupta2018,Chandra2019,Deo2018} have been studied separately. As a result, modern perception systems often perform 3D MOT and forecasting separately in a cascaded order, where tracking is performed first to obtain trajectories in the past, followed by trajectory forecasting to predict trajectories in the future. However, this cascaded pipeline with separately trained modules can lead to sub-optimal performance, as information is not shared during training. Since tracking and forecasting modules are highly dependent, it would be beneficial to optimize them jointly. For example, a better MOT module can lead to better performance of its downstream forecasting module while a more accurate motion model learned in trajectory forecasting can improve data association for MOT.

\vspace{-0.4cm}
\section{A Joint 3D MOT and Trajectory Forecasting Model}
\vspace{-0.2cm}

% Solution statement
Our goal is to jointly optimize the MOT and forecasting modules and enable the propagation of performance information through the entire system during training. Instead of running two modules separately in a sequential order as shown in Fig.~\ref{fig:teaser} (top), we propose to perform MOT and forecasting in parallel as shown in Fig.~\ref{fig:teaser} (bottom). As a result, the gradients computed in both heads (one for each task) can be propagated back to learn a better shared feature representation for both tasks. By keeping the MOT and forecasting heads parallel, i.e., forecasting does not explicitly depend on the MOT results, we can prevent association errors made in MOT from directly influencing the forecasting module. The forecasting module can still use the implicit MOT information encoded in the shared features computed by the GNN.

\vspace{-0.5cm}
\section{Social Interaction Modeling with Graph Networks}
\vspace{-0.25cm}

% Motivation for GNN
Modeling interaction for 3D MOT is crucial in the presence of multiple agents but has often been overlooked in prior work. Prior work in 3D MOT extracts the feature of each object \textit{independently}, i.e., each object's feature only depends on the object's own inputs (image crop or location). As a result, there is no interaction between objects. We found that \textit{independent feature extraction leads to inferior discriminative feature learning}, and object dependency is the key to obtaining discriminative features. Intuitively, the features of the same object over two frames should be as similar as possible and the features between two different objects should be as different as possible to avoid confusion during data association. This can only be achieved if object features can be obtained as a dependent or context-aware process, i.e., modeling interactions between objects.

% Solution statement
To model interaction in 3D MOT, we employ a feature interaction mechanism as shown in Fig.~\ref{fig:teaser} (Bottom) by introducing Graph Neural Networks (GNNs) to 3D MOT. Specifically, we construct a graph with each node being an object in the scene. Then, at every layer of the GNNs, each node can update its feature by aggregating features from other nodes. This node feature aggregation is useful because the resulting object features are no longer isolated and are adapted according to other objects. We observe in our experiments that, after a few GNN layers, the affinity matrix becomes more discriminative than the affinity matrix obtained without interaction. In addition to using GNNs to model interaction for 3D MOT, GNNs can also provide socially-aware context to improve trajectory forecasting \cite{Chandra2019_2}. To the best of our knowledge, we are the first to employ GNNs in a unified 3D MOT and trajectory forecasting method. 

\vspace{-0.5cm}
\section{Diversity Sampling for Trajectory Forecasting}
\vspace{-0.25cm}

% Motivation for diversity sampling
Since future trajectories of objects should be stochastic and multi-modal due to many unobserved factors (e.g., hidden intentions), prior work in trajectory forecasting often learns the future trajectory distribution with deep generative models. At test time, these methods randomly sample a set of future trajectories from the generative model without considering the correlation between samples. As a result, the samples can be very similar and only cover a limited number of modes, leading to poor sample efficiency. This inefficient sampling strategy is harmful in real-time applications because producing a large number of samples can be computationally expensive and lead to high latency. Moreover, without covering all the modes in the trajectory distribution and considering all possible futures, the perception system cannot plan safely, which is important in safety-critical applications such as autonomous driving.

% Solution statement
To improve sample efficiency in trajectory forecasting, we depart from the random sampling in prior work and employ a diversity sampling technique that can generate diverse trajectory samples from a pretrained CVAE model. The idea is to learn a separate sampling network which maps each object's feature to a set of latent codes. The latent codes are then decoded into trajectory samples. In this way, the produced samples are correlated (unlike random sampling where the samples are independent), which allows us to enforce structural constraints such as diversity onto the samples. Specifically, we use determinantal point processes (DPPs) to optimize the diversity of the samples. 

% ==================================================================================================================

\vspace{-0.5cm}
\section{Experiments}
\vspace{-0.25cm}

\textbf{Datasets}. We use standard autonomous driving datasets: KITTI \cite{Geiger2012}. Also, since there is no existing evaluation procedure that can jointly evaluate 3D MOT and trajectory forecasting, we evaluate two modules separately and compare against prior work on each individual module of our joint method. For KITTI, same as most prior works, we report results on the car subset for comparison. 

% ==================================================================================================================

% \vspace{-0.4cm}
% \subsection{Evaluating 3D Multi-Object Tracking}
% \vspace{-0.1cm}

\begin{table}[t]
\vspace{-0.5cm}
\caption{3D MOT evaluation on the KITTI dataset.}
\centering
\resizebox{\textwidth}{!}{
\begin{tabular}{@{}lrrrrrrr@{}}
\toprule
Methods \ \ \ \ \ \ \ \ & \ \ \textbf{sAMOTA}(\%)$\uparrow$ & \ \ AMOTA(\%)$\uparrow$ & \ \ \ AMOTP(\%)$\uparrow$ & \ \ \ MOTA(\%)$\uparrow$ & \ \ \ MOTP(\%)$\uparrow$ & \ \ \ IDS$\downarrow$ & \ \ \ FRAG$\downarrow$ \\
\midrule
FANTrack~\cite{Baser2019}       & 82.97 & 40.03 & 75.01 & 74.30 & 75.24 & 35 & 202\\
AB3DMOT\cite{Weng2019}          & 93.28 & 45.43 & \textbf{77.41} & 86.24 & \textbf{78.43} & \textbf{0} & 15\\
\midrule
\textbf{Ours}                   & \textbf{94.41} & \textbf{46.15} & 76.83 & \textbf{86.89} & 78.32 & 3 & 8 \\
\bottomrule
\end{tabular}}
\vspace{-0.7cm}
\label{tab:3dmot_quan}
\end{table}

\vspace{1.5mm}\noindent\textbf{Evaluating 3D Multi-Object Tracking.} We use standard CLEAR metrics (including MOTA, MOTP, IDS) and new sAMOTA, AMOTA and AMOTP metrics \cite{Weng2019} for evaluation. We summarize the results in Table \ref{tab:3dmot_quan}. Our method consistently outperforms baselines in sAMOTA, which is the primary metrics for ranking MOT methods. We hypothesize that this is because our method leveraging GNN obtains more discriminative features to avoid confusion in MOT association while all 3D MOT baselines ignore the interaction between objects. Moreover, joint optimization of the tracking and forecasting modules might help. 

\begin{table}[t]
\begin{center}
\vspace{-0.5cm}
\caption{Trajectory forecasting evaluation on the KITTI dataset.}
\resizebox{\textwidth}{!}{
\begin{tabular}{@{}llrrrrrrr@{}}
\toprule
Settings \ \ \ \ \ \ \ \ \ & Metrics \ \ & \ \ Conv-Social \cite{Deo2018} & \ \ Social-GAN \cite{Gupta2018} & \ \ TraPHic \cite{Chandra2019} & \ \ Graph-LSTM \cite{Chandra2019_2} & \ \ \ \ \ \ \ \ \textbf{Ours} \\ 
\midrule

\multirow{4}{*}{KITTI-1.0s} & ADE$\downarrow$ & 0.607 & 0.586 & 0.542 & 0.478 & \textbf{0.471}\\
                            & FDE$\downarrow$ & 0.948 & 1.167 & 0.839 & 0.800 & \textbf{0.763}\\ 
                            & ASD$\uparrow$   & 1.785 & 0.495 & 1.787 & 1.070 & \textbf{2.351}\\ 
                            & FSD$\uparrow$   & 1.987 & 0.844 & 1.988 & 1.836 & \textbf{4.071}\\ 
\midrule

\multirow{4}{*}{KITTI-3.0s} & ADE$\downarrow$ & 2.362 & 2.340 & 2.279 & 1.994 & \textbf{1.319}\\
                            & FDE$\downarrow$ & 3.916 & 4.102 & 3.780 & 3.351 & \textbf{2.299}\\ 
                            & ASD$\uparrow$   & 2.436 & 1.351 & 2.434 & 2.745 & \textbf{5.843}\\ 
                            & FSD$\uparrow$   & 2.973 & 2.066 & 2.973 & 4.582 & \textbf{10.123}\\ 
\bottomrule
\end{tabular}
\label{tab:trajectory_quan} 
}
\end{center}
\vspace{-1cm}
\end{table}

\vspace{1.5mm}\noindent\textbf{Evaluating Trajectory Forecasting.} We use the standard metrics: Average Displacement Error (ADE) and Final Displacement Error (FDE) for evaluation. Additionally, to evaluate the diversity of the trajectory samples, we use Average Self Distance (ASD) and Final Self Distance (FSD) in \cite{yuan2019diverse} for sample diversity evaluation. We summarize the results in Table \ref{tab:trajectory_quan}. Our method, which (1) is jointly trained with a 3D MOT head, (2) uses GNNs for feature interaction and (3) uses diversity sampling, outperforms the baselines in both accuracy and diversity metrics. Particularly, our method outperforms baselines for the long-horizon (i.e., 3.0s) experiment. This is because our method has a higher sample efficiency and can cover different modes of the future trajectory distribution. 

\vspace{-0.4cm}

\bibliographystyle{splncs04}
\bibliography{main}

\end{document}